\documentclass{crorr}

\usepackage[labelfont={sf}]{caption}
\usepackage{graphicx}
\usepackage{subcaption}
\usepackage{float}
\usepackage{algorithmic}
\usepackage{algorithm}
\usepackage{epstopdf}
\usepackage{comment}
\usepackage{color}
\usepackage{tabularx}
\usepackage{biblatex}

\usepackage[colorlinks=true,urlcolor=black,citecolor=blue]{hyperref}

\renewcommand{\rec}{}

\makeatletter
\def\ps@firstpage{
  \let\@oddhead\@empty
  \let\@evenhead\@empty
  \let\@oddfoot\@empty
  \let\@evenfoot\@empty
}
\makeatother

\begin{document}

	\title{Comparative Analysis of Modern Machine Learning Models for Retail Sales Forecasting}

	\author[Hobor, Brčić, Polutnik, Kapetanović]{\bf Luka Hobor\affil{1, 2}\corrauth, \,  Mario Brčić\affil{1, 2}, \, Lidija Polutnik\affil{3} and Ante Kapetanović\affil{4}}
	\address{\affilnum{1}\ Faculty of Electrical Engineering and Computing,  University of Zagreb, Unska 3 Zagreb, Croatia  \\
		E-mail: $\langle$\{luka.hobor, mario.brcic\}@fer.hr$\rangle$
		\\
		\medskip
		\affilnum{2}\ It From Bit d.o.o., Zagreb, Croatia \\ 
        \affilnum{3}\ Babson College, Babson Park, Massachusetts, United States \\ 
		E-mail: $\langle$polutnik@babson.edu$\rangle$
        \affilnum{4}\ mStart Plus d.o.o., Zagreb, Croatia \\ 
		E-mail: $\langle$Ante.Kapetanovic@mstart.hr$\rangle$
	}

	\begin{abstract}
		Accurate demand forecasting is critical for brick-and-mortar retailers to optimize inventory management and minimize costs. This study evaluates statistical baselines, tree-based ensembles (XGBoost and LightGBM), and deep learning architectures (N-BEATS, N-HiTS, and the Temporal Fusion Transformer) on retail sales data characterized by intermittent demand, substantial missingness, and frequent product turnover. Models are compared across four configurations varying by aggregation level and imputation strategy, using evaluation protocols that reflect typical deployment patterns for each model class. Localized tree-based methods achieve superior performance, with XGBoost attaining the lowest RMSE of 4.833. While SAITS-based imputation improved neural network performance in aggregated settings, these models remained inferior to ensemble methods. The results suggest that, under the studied constraints, model selection should prioritize alignment with problem characteristics over architectural sophistication. 

	\end{abstract}
	\keywords{gradient-boosted decision trees, neural networks, predictive analytics, retail sales forecasting, time-series analysis}
	\maketitle
	\bigskip
	\noindent
\section{Introduction}

Accurate sales forecasting is indispensable in retail, enabling better inventory planning, resource allocation, and the achievement of  revenue and profit goals. The choice of the best possible forecasting model to use is essential as retailers aim to gain a competitive advantage in the context of irregular demand and limited historical sales data in various product categories. While classical approaches such as the Exponential Time Smoothing \cite{winters1960forecasting} or the Theta model \cite{ASSIMAKOPOULOS2000521} have historically formed the backbone of retail forecasting, they often fall short in capturing modern retail complexities and operational nuances. Recent advances in machine learning and deep learning present significant opportunities for retailers to improve their sales predictions and operational efficiencies with new alternatives, such as tree-based models like XGBoost \cite{xgboost} and LightGBM \cite{lightgbm}, as well as neural architectures such as N-BEATS \cite{nbeats}, NHITS \cite{nhits}, TFT \cite{tft}, and others \cite{patchTST, imputeFormer, moderntcn}.

Recent comparative evaluations of forecasting models have shown mixed results across domains. Studies such as \cite{mcelfresh2024neuralnetsoutperformboosted, Ramesh_Usman_2024, Borisov, chen2024excelformerneuralnetworksurpassing} have explored the performance of neural networks versus gradient boosting models and highlighted that deep learning does not consistently outperform tree-based approaches, especially on tabular, sparse, or highly intermittent retail data. These findings align with results from the M5 forecasting competition \cite{m5_summary}, which used Walmart’s highly granular retail sales data consisting of daily item-store records. While some series exhibited intermittent demand patterns, the dataset primarily featured dense, continuous observations with limited explicit missingness. Despite this, LightGBM-based ensembles outperformed deep learning models, including those developed by Amazon’s forecasting team \cite{amazonM5}. On the other hand, recent findings from Zalando suggest that transformer-based models exhibit scaling laws in retail forecasting: as the volume of training data increases, demand forecasting error decreases in a predictable manner \cite{kunz2023deeplearningbasedforecasting}. These understandings motivate a closer investigation into the conditions under which different forecasting paradigms perform best.

Much of the recent deep learning success has been in large-scale online retail settings, where companies like Amazon \cite{amazon_forecasting_2021} and Zalando \cite{kunz2023deeplearningbasedforecasting} operate centralized warehouses and enjoy the benefits of aggregated demand and operational homogeneity. In contrast, brick-and-mortar (B\&M) retail is much more fragmented: sales occur in thousands of physical stores, each with limited shelf space, store-level variability, and frequent changes in product assortment. These conditions lead to noisy, intermittent demand signals that are known to affect neural forecasting models\cite{boylan2021intermittent, KOURENTZES2013198, WEERAKODY2021161}. 

In this study, we conducted a large-scale evaluation of forecasting models in the context of B\&M retail, using real-world data from a major retailer in Southeast Europe (SEE). Our primary goal was to systematically compare the performance of diverse modeling approaches, including statistical baselines, tree-based ensembles, and state-of-the-art neural architectures, in providing daily sales predictions for a horizon of 26 weeks (182 days) into the future for products within the hygiene category. A key dimension of our comparison involves contrasting local modeling strategies, where separate models are trained for individual product groups, against global approaches that train a single model across all product groups jointly. This daily forecasting resolution was strategically chosen to capture swift changes in sales often driven by intra-week promotional activities and pricing changes, which occur within the retailer's planning cycle. For evaluation purposes, the daily predictions were subsequently aggregated to a weekly basis to assess forecast accuracy across the 26-week horizon.

The models deployed by the research team are benchmarked for their ability to handle operational realities of physical retail, including intermittent demand, missing values, and product censorship due to assortment shifts and other changes in the retail environment. Our work builds on lessons from academic literature and industrial benchmarks, including the Rossmann case study \cite{Middel_Davel} and the M5 competition \cite{m5_summary}, and provides new insights into model performance evaluation under real-world retail constraints.

Our main contributions are threefold: we offer practical guidance for retail practitioners through an end-to-end modeling pipeline for long-horizon retail forecasting encompassing data preprocessing, feature engineering, and model training strategies; we provide comprehensive empirical comparisons of state-of-the-art forecasting models across multiple dimensions including model architecture (statistical, tree-based, neural), modeling scope (local vs. global), and data preprocessing strategies (imputed vs. nonimputed); and we demonstrate that localized tree-based models consistently outperform global neural architectures in brick-and-mortar retail settings while highlighting the limitations of neural models under operational constraints typical of physical retail.

\section{Methodology and Data}

\subsection{Data Description}

The dataset includes daily retail sales information with multiple dimensions, including time of transaction, store characteristics, product attributes, prices of sold products, promotional activities, and inventory changes. Each observation represents a daily record for a specific product-store combination, making the dataset well suited for longitudinal analysis across multiple hierarchical levels.

The area studied consists of geographies that have different levels of importance in terms of their contributions to revenue, market size, and overall positioning. The data provided includes the identification of the zones for the purpose of business strategy. Products themselves follow a three-level hierarchy: individual items are first organized into categories based on their main purpose, then into subcategories called groups and within each group, related products are further clustered into units-of-need (UoNs). Our dataset consisted of one category for hygiene products, which contained 27 groups and 79 UoNs in total over 446 stores.

Other product metadata includes unit pricing (with and without tax), cost of goods sold and promotions, which are extensively captured through binary and count features indicating tactical promotions and various loyalty initiatives. Although inventory data are fully tracked, including physical stock levels, incoming deliveries, and reserved or frozen quantities, we recognize that inventory record inaccuracy remains a well-documented issue in retail systems\cite{rekik2025inventoryrecordinaccuracygrocery}.

\begin{table}[h!]
\centering
\begin{tabular}{l r}
\hline
\textbf{Metric} & \textbf{Value} \\
\hline
Series Count & 46841 \\
Missingness & 20.933\% \\
Coverage Ratio & 78.848\% \\
Mean Series Length & 726.923 \\
Median Series Length & 794 \\
Minimum Series Length & 2 \\
Maximum Series Length & 1124 \\
Mean Zero-Sale Days within Series & 491.801 \\
Percent of Series with at least One Zero-Sale Day & 99.981\% \\
\hline
\end{tabular}
\caption{Time Series Descriptive Statistics}
\end{table}

Table 1 summarizes the dataset characteristics. The dataset comprises 46,841 daily SKU-level time series with a mean length of 727 observations and a median of 794. The missingness rate of 20.9\% and coverage ratio of 78.8\% reflect typical retail operations. Nearly all series (99.9\%) contain at least one zero-sale day, with an average of 492 zero-sale days per series, indicating highly intermittent demand patterns. These characteristics pose challenges for traditional forecasting approaches designed for continuous demand patterns.

The dataset structure and challenges are comparable to those reported in major demand forecasting benchmarks such as the M5 Competition\cite{m5_summary} or Rossmann study case \cite{Middel_Davel}. However, unlike e-commerce datasets, which benefit from denser demand signals, B\&M retail environments are more exposed to localized promotions and physical differences across stores.


\subsection{Data Preprocessing and Feature Selection}

To ensure high-quality inputs for forecasting, a structured data preprocessing pipeline was implemented, including enrichment, imputation, transformation, and formatting steps appropriately adjusted for time series models.

\textbf{Data Loading and Standardization.} The preprocessing pipeline begins by loading raw retail data and applying systematic column renaming and standardization to ensure consistency. Invalid entries are cleaned, and leaky or unstable columns, particularly inventory bookkeeping fields, are removed to prevent data leakage. Time-based and categorical features are constructed, including calendar effects and indicators for holidays, stock availability, promotional activity, and pension-day flags, the latter being particularly relevant for understanding demand patterns in retail contexts. A critical step in this phase involves discarding products that were unavailable or out of stock during the observation period. This filtering eliminates out-of-stock bias, which is essential for sales forecasting problems \cite{hyndman2018forecasting, chopra2019supply}, as it ensures the model learns from genuine demand patterns rather than supply constraints. The target variable is clipped and sanitized to remove outliers and implausible values.

\textbf{Competitive and Macroeconomic Environment.} The point-of-sale, store-specific data were augmented with relevant competitive and macroeconomic indicator data. In order to control for the level of competition in the relevant geographic market, the count of competitors within a 1-kilometer radius and their prices  for comparable items were added to the dataset. Further, macroeconomic environment variables from the National Statistical Office, such as Consumer Price Index (CPI), average salaries (national and regional), and population estimates within specific geographies, were  included in order to control for the purchasing power and the size of the market. 

\textbf{Handling Missing Values.} Our dataset is structured to include an entry for every SKU on every day when it is supposed to be in stores, regardless of whether a sale occurred. This means that a zero value explicitly indicates no sales, while missing values represent genuinely unavailable data for auxiliary variables (e.g., competitor prices), not the sales themselves. The pipeline addresses missing covariate values through a multi-stage approach: horizontal imputation across related features, forward and backward fill for time-continuous numeric variables, and group-level imputation based on store or item characteristics. These strategies ensure temporal consistency and are only applied to static features(e.g., item metadata) or ones that are rarely changed or occur in a predictable manner (e.g., close-date prices, costs, macro indicators).

To explore whether more sophisticated methods could enhance model performance, we conducted a separate experiment using a deep learning imputation model. Specifically, we trained and applied the SAITS model \cite{saits}, implemented via the PyPOTS library \cite{du2023pypots}, to impute missing values for the target and other variables in the training and validation sets.

\begin{figure}[ht!]
\centering\includegraphics[width=0.8\linewidth]{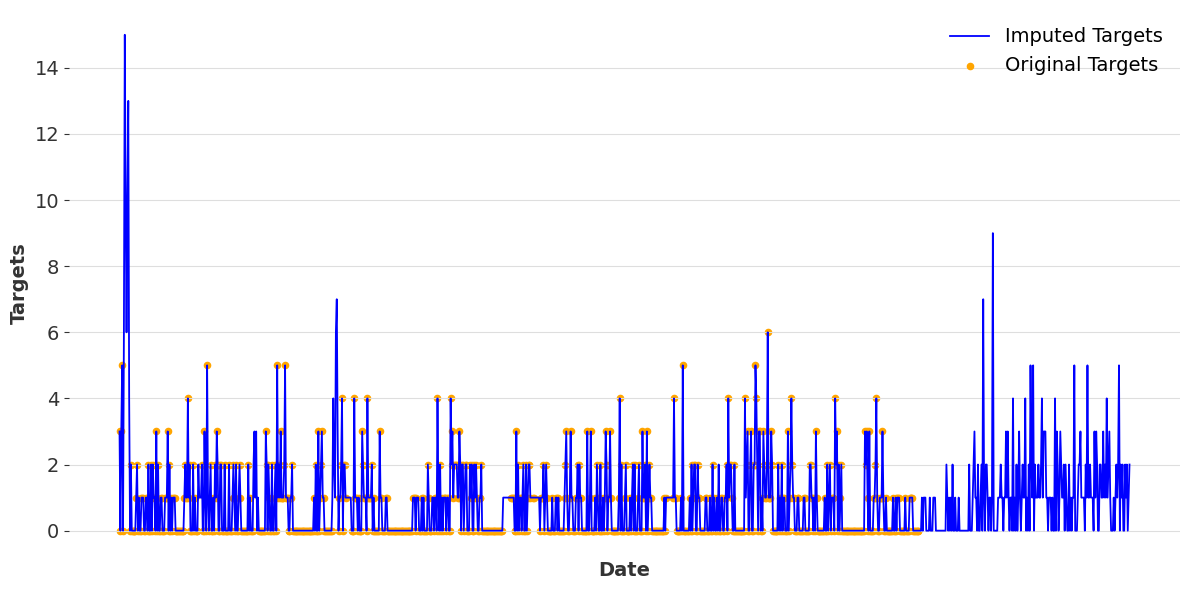}
\caption{Sales Values Real and Imputed}
\end{figure}

To rigorously assess imputation quality, we conducted a three-part analysis comparing SAITS with simpler alternatives. First, we examined distributional differences between original and imputed values. SAITS-imputed values exhibit significantly different distributions from originals (all Kolmogorov-Smirnov p-values $\approx$ 0), with compressed variance and capped maximum values. For example, in product one group, imputed values show a standard deviation of 0.53 versus 5.42 for originals, with maximum values of 23 versus 353.

Second, we compared SAITS against simpler imputation methods including zero-fill, forward-fill, linear interpolation, and group-median imputation. Linear interpolation performs best for higher-volume groups, while zero-fill and group-median perform better for sparse groups. SAITS produces values weakly correlated with simple methods ($|r| < 0.15$), suggesting it captures different underlying patterns.

Finally, we conducted a LightGBM deep dive on a random group. Top-3 feature rankings remain stable regardless of imputed data inclusion.

\textbf{Feature Engineering.} Several engineered features were introduced to enrich the temporal signal and provide the models with interpretable and context-aware inputs. The pipeline first trims observations after a specified cutoff date to define the modeling window. Lag and rolling-window statistics were computed for key variables such as sales, prices, and promotional activity using rolling, expanding, and exponentially weighted windows across multiple aggregation periods, specifically weekly, monthly, and yearly. The data are restructured by temporarily indexing on timestamps, and indicators are generated for whether a store was operating on the previous and next days to capture operational continuity patterns.

In-store relative positioning of products within the same Unit of Need (UoN) was captured through metrics like minimum and maximum price among comparable SKUs. All nominal prices were converted into real prices using CPI-adjusted values to control for inflation. Baseline prices for each item were established by extracting standard zone reference prices, filling them across stores, and adjusting for inflation. From these baselines, pricing-strategy features were derived, including promotion discounts, price multipliers, and baseline-relative indices. Relative pricing indicators were also constructed to measure the retailer's price advantage or disadvantage compared to competitors (out of store).

To prevent data leakage and ensure realistic forecasting conditions, all covariates were explicitly classified as either past-only or future-known variables. Past-only covariates include variables that are only observable up to the forecast origin and cannot be reliably predicted into the future: historical sales and derived features (lags, rolling statistics), competitor prices and counts (observable only historically), stock availability flags, etc. Future-known covariates include variables that are either constant, deterministic, or under direct control of the retailer over the forecast horizon: calendar features (day of week, month, holidays, pension days), store characteristics and metadata (location, size, zone), CPI projections (using forward-filled values from the most recent observation or exponential time smoothing for longer periods), planned promotional flags (directly controlled by the scenario simulation) and baseline and planned prices (inflation-adjusted, under the simulator's control).

For variables that exhibit both historical variability and forward-looking characteristics, such as competitor prices and certain inventory-related features, we applied controlled noise injection to smooth potential forward-looking signals that might leak into the training data. Specifically, competitor prices were treated as past-only during forecast generation, with no assumptions made about their future values. The pipeline enforces consistent item and store metadata throughout the temporal window. Revenue gaps identified in the historical period were repaired using median-based imputation, but no imputation was applied to future periods where values are genuinely unknown. This explicit separation ensures that both tree-based ensemble models and neural network architectures are trained and evaluated under identical, leakage-free conditions.

\textbf{Feature Selection.} Feature selection was conducted in a single round early in the development process and remained fixed afterward. Core variables grounded in economic theory, such as prices, promotions, calendar effects, and competitor signals, were retained by default. To reduce model complexity and enhance generalization, a filtering procedure was applied to additional engineered features based on their predictive utility. We used the Boruta algorithm \cite{boruta}, with LightGBM as the background estimator, to identify and keep only features with statistically verified predictive relevance. Features with excessive missingness, low variance, or multicollinearity were discarded to enhance efficiency and stability.

\textbf{Train-Validation Split.} A time-based cutoff was used to divide the dataset into training and validation periods, and all series had observations in both periods. 

\textbf{Categorical Encoding.} Categorical variables were encoded using the CatBoost encoder \cite{catboost} from the package \texttt{category\_encoders}.

\textbf{Final Time-Series Transformation.} For neural forecasting models, data were organized into time-series objects by grouping over product and store (SKU) identifiers. Target and numerical covariates were normalized, and to ensure series continuity, missing data points were simply interpolated and endpoints were replaced with forward and backward fills. Appropriate masks were created so that models are not trained on these points. For scalability sake, numerical fields were downcast to lighter data types to optimize memory usage during model training on large datasets.

\subsection{Overview of Forecasting Models and Implementation}

This study evaluates the performance of a range of forecasting models commonly applied in time-series prediction. These models include tree-based ensembles, neural networks, and classical statistical approaches. 

All neural network models were implemented using the \texttt{Darts} library, which supports modular time-series architectures and standardized training workflows. Each model was trained across four experimental setups: (1) training one model for each product group independently using nonimputed data; (2) training on each product group using imputed data; (3) training across all product groups jointly on nonimputed data; and (4) training jointly using the imputed dataset. This design allowed us to assess the impacts of SAITS-based imputation and group-level modeling on sales forecasting results.

To optimize model performance, a hyperparameter search was conducted using the HEBO algorithm \cite{hebo}, a scalable Bayesian optimization method. Due to computational constraints, hyperparameter tuning was conducted separately on a randomly sampled 10\% subset for both nonimputed and imputed data configurations. This ensured that each experimental condition received appropriately optimized hyperparameters rather than applying configurations tuned on one data type to another.

\subsubsection{Training and Evaluation}

The models were primarily trained using a Poisson likelihood objective to predict daily sales volumes, with the exception of the Temporal Fusion Transformer model, which was optimized using a Gaussian likelihood (a configuration determined through HEBO).


\textbf{Forecasting Strategy:} Two evaluation protocols were employed, each reflecting standard deployment practices for its respective model class. For tree-based models, a one-step-ahead rolling protocol was used: for each day $t$, the model receives all information available up to $t$ and predicts day $t+1$. This is consistent with their treatment of each observation independently and their widespread success in competitions\cite{m5_summary, amazonM5}. Neural models use a direct multi-horizon strategy, generating all 182 daily forecasts in a single forward pass, consistent with how architectures like N-BEATS, NHiTS, etc. are designed to learn sequential dependencies and cross-horizon patterns. To verify that this protocol difference did not disadvantage neural models, we also evaluated neural networks using rolling-origin prediction, matching the tree-based protocol. Results showed that neural networks with rolling-origin evaluation performed equally or worse than fixed-origin multi-horizon prediction, confirming that the observed performance differences reflect genuine model capabilities rather than evaluation protocol asymmetry.

\textbf{Forecast Horizon and Aggregation:} Each model generated one sales forecast per SKU per day over a 26-week horizon. This daily resolution was intentionally chosen to accurately capture high-frequency sales volatility resulting from intra-week promotional and pricing changes that align with the retailer's operational planning cycle. The daily forecasts were subsequently aggregated to weekly totals, and performance metrics were computed at this weekly level to facilitate interpretation and align with business planning requirements.

In addition to standard forecasting evaluation metrics, we measured financial performance by computing demand error and bias \cite{kunz2023deeplearningbasedforecasting}:

$$
D_{T,h} = \frac{\sum_{i} \sum_{T=t+1}^{t+h} b_i (\hat{q}_{i,T} - q_{i,T})^2}{\sum_{i} \sum_{T=t+1}^{t+h} b_i q_{i,T}^2} 
$$

$$
B_{T,h} = \frac{\sum_{i} \sum_{T=t+1}^{t+h} b_i (\hat{q}_{i,T} - q_{i,T})}{\sum_{i} \sum_{T=t+1}^{t+h} b_i q_{i,T}} 
$$

where $b_i$ is the price of SKU and $\hat{q}_{i,T}, q_{i,T}$ are the predicted and actual sales values at a certain date.

For experiments involving imputed data, evaluation was performed exclusively on the original (nonimputed) validation data points to ensure a fair comparison across models.

All experiments were executed on a workstation equipped with an AMD Ryzen Threadripper PRO 7985WX 64-core CPU, 256 GB of RAM, and an NVIDIA RTX 4080 GPUs.

\section{Results and Discussion}
\subsection{Results}

\begin{table}[ht]
\centering
\footnotesize
\begin{tabularx}{\textwidth}{X | X | X | X | X | X | X | X | X | X | X | X | X | X | X | X}
\hline
Model & RMSE & MAE & Demand Error & Demand Bias & RMSSE & MASE & ME & RB \\
\hline
\multicolumn{5}{l}{\textbf{Baseline}} \\
\hline
Mean & 12.022 & 4.947 &  1.211 & \textbf{-0.028} & 1.476 & 1.871 & -0.500 & -0.136\\
Theta & 27.128 & 6.779 & 2.854 & -0.356 & 3.331 & 2.563 & -1.610 & -0.440\\
ETS & 26.027 & 6.355 & 2.585 & -0.291 & 4.369 & 2.403 & -1.331 & -0.367\\
CSBA & 34.972 & 7.989 & 1.072 & 0.084 & 1.795 & 3.648 & 0.419 & 0.314\\
\hline

\multicolumn{5}{l}{\textbf{Case A: Individual Groups}} \\
\hline
LGBM & \underline{4.847} & \underline{1.958}  & 0.484 & \underline{-0.040} & \underline{0.595} & \underline{0.740} & \underline{-0.205} & \underline{-0.056} \\
NB & 13.629 & 5.132 & 1.133 & 0.775 & 1.482 & 1.729 & 3.270 & 0.736 \\
NH & 12.764 & 4.575 & 1.254 & 0.948 & 1.537 & 1.712 & 3.252 & 0.849 \\
TFT & 7.773 & 3.049 & 0.753 & 0.220 & 0.945 & 1.130 & 0.673 & 0.175 \\
XGB & \textbf{4.833} & \textbf{1.935} & \textbf{0.481} & -0.085 & \textbf{0.593} & \textbf{0.732} & -0.356 & -0.097 \\
\hline
\multicolumn{5}{l}{\textbf{Case B: Whole Category}} \\
\hline
LGBM & 4.949 & 2.033 & \underline{0.483} & -0.105 & 0.601 & 0.754 & -0.450 & -0.117 \\
NB & 13.744 & 6.018 & 1.345 & 0.387 & 1.670 & 2.231 & 0.910 & 0.236 \\
NH & 13.744 & 6.018 & 1.345 & 0.387 & 1.670 & 2.231 & 0.910 & 0.236 \\
TFT & 14.060 & 6.087 & 1.357 & 0.398 & 1.708 & 2.257 & 1.130 & 0.294 \\
XGB & 5.045 & 2.041  & 0.493 & -0.122 & 0.613 & 0.757 & -0.506 & -0.132 \\
\hline
\multicolumn{5}{l}{\textbf{Case C: Individual Groups, Imputed Train Data}} \\
\hline
LGBM & 10.758 & 5.036  & 1.117 & 0.922 & 1.307 & 1.867 & 2.578 & 0.686 \\
NB & 15.362 & 7.704 & 0.628 & 0.065 & 1.867 & 2.856 & 1.210 & 0.085 \\
NH & 16.352 & 7.980 & 0.675 & 0.122 & 1.987 & 2.959 & 1.846 & 0.130 \\
TFT & 15.579 & 8.274 & 0.626 & 0.074 & 1.893 & 3.068 & 1.265 & 0.089 \\
XGB & 6.027 & 2.310  & 0.616 & -0.224 & 0.732 & 0.856 & -0.741 & -0.197 \\
\hline
\multicolumn{5}{l}{\textbf{Case D: Whole Category, Imputed Train Data}} \\
\hline
LGBM & 7.932 & 3.450 & 0.769 & -0.053 & 0.964 & 1.279 & \textbf{-0.077} & \textbf{-0.020} \\
NB & 7.853 & 2.928 & 0.773 & 0.067 & 0.954 & 1.086 & 0.263 & 0.069 \\
NH & 7.789 & 2.894 & 0.773 & 0.069 & 0.946 & 1.073 & 0.272 & 0.072 \\
TFT & 7.134 & 2.971 & 0.686 & 0.130 & 0.867 & 1.102 & 0.516 & 0.137 \\
XGB & 6.107 & 2.349 & 0.620 & -0.307 & 0.742 & 0.871 & -1.106 & -0.294 \\
\hline
\end{tabularx}
\caption{Summary Statistics. \scriptsize{Abbreviations: ETS: Exponential Time Smoothing; CSBA: Croston SBA; TFT: Temporal Fusion Transformer; NH: NHiTS; NB: N-BEATS; LGBM: LightGBM; XGB: XGBoost; RB: Relative Bias; Best Results are Bolded and Second Best Underlined}}
\end{table}

The results demonstrate clear superiority of gradient boosting methods (XGBoost and LightGBM) over deep learning approaches across most scenarios. In Case A (Individual Groups), XGBoost achieves the best performance with an RMSE of 4.833 and an MAE of 1.935, closely followed by LightGBM. The scale-independent metrics (RMSSE of 0.593 and MASE of 0.732) indicate significant error reduction compared to naive baselines. This pattern suggests that tree-based models' ability to capture complex feature interactions and handle heterogeneous data provides significant advantages for this retail forecasting problem.

Comparing Case A to Case B reveals minimal performance degradation for gradient boosting models, with XGBoost and LightGBM maintaining strong performance. However, deep learning models show deterioration, especially TFT, suggesting difficulty in capturing diverse patterns across the broader category scope. The introduction of SAITS-based imputation (Cases C and D) produces mixed results. In Case C, LightGBM experiences dramatic performance degradation (RMSE increases from 4.847 to 10.758), while XGBoost demonstrates greater resilience (RMSE of 6.027), possibly reflecting its superior regularization strategies. Case D shows improved performance over Case C for almost all models, with deep learning approaches becoming competitive, though XGBoost still maintains an edge.

The bias metrics reveal important systematic tendencies. Ensemble-based models exhibit slight negative bias (underestimation), while neural networks exhibit systematic overestimation. Case D achieves the most balanced predictions across models, with LGBM showing minimal bias (ME of -0.077, RB of -0.020). The imputation process significantly affects bias patterns, with Case C showing LightGBM shifting to severe positive bias (Demand Bias of 0.922), while Case D demonstrates that category-level imputation better preserves bias characteristics.

\begin{figure}[ht!]
\centering\includegraphics[width=0.9\linewidth]{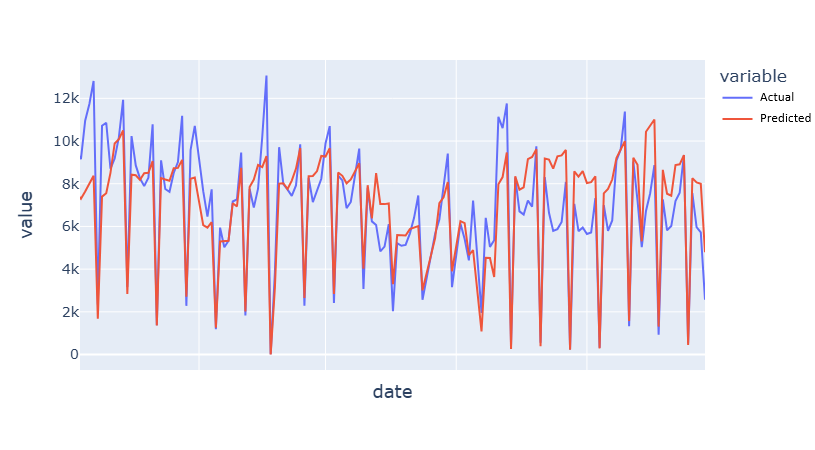}
\caption{Example of Aggregated Predicted vs Actual Sales on a Single Group}
\end{figure}

\begin{figure}[ht!]
 \centering
 \begin{subfigure}{0.45\linewidth}
  \centering
  \includegraphics[width=\linewidth]{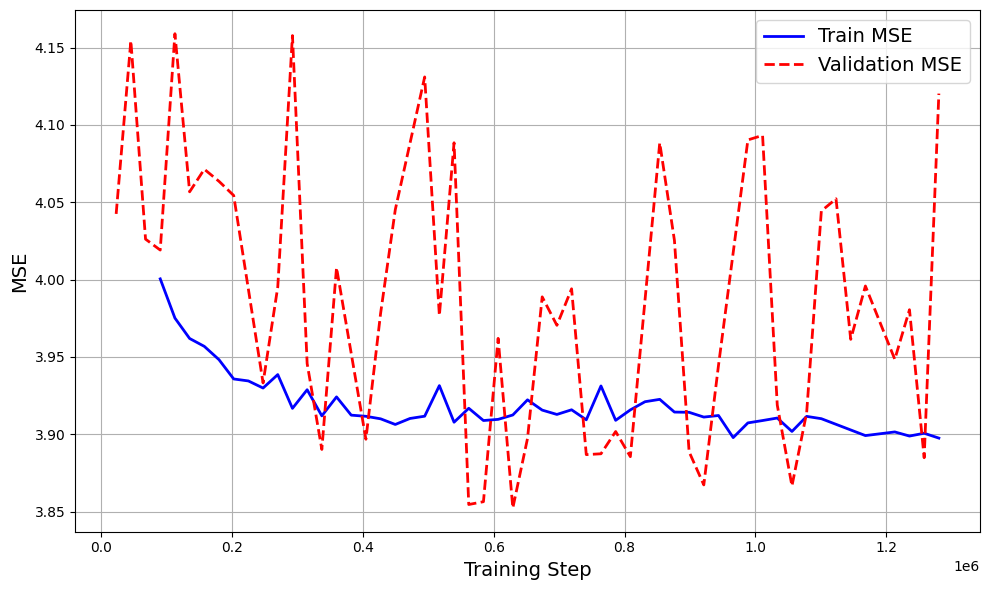}
  \caption{N-BEATS (WC, imp): Train vs. Validation MSE}
 \end{subfigure}
 \hfill
 \begin{subfigure}{0.45\linewidth}
  \centering
  \includegraphics[width=\linewidth]{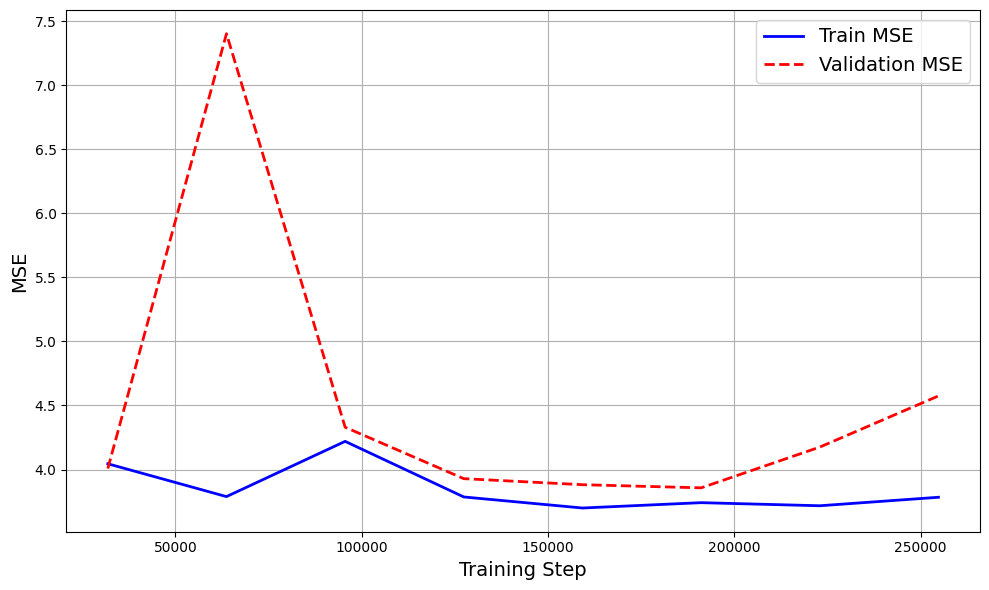}
  \caption{TFT (WC, raw): Train vs. Validation MSE}
 \end{subfigure}

 \caption{Comparison of Training and Validation MSE for N-BEATS and TFT Models.} 
\end{figure}

\begin{table}[ht]
\centering
\begin{tabular}{lrrrrr}
\hline
 & N-BEATS & NHiTS & TFT & LGBM & XGB \\
\hline
Training Time & 146.715 & 145.928 & 4424.972 & 35.566 & 14.542 \\
\hline
\end{tabular}

\caption{Training Time Statistics (in Minutes) \footnotesize{Per Epoch for Neural Networks, Total for LGBM and XGB}}
\end{table}

\begin{table}[ht]
\centering
\footnotesize
\begin{tabularx}{\textwidth}{l | X | X | X}
\hline
\textbf{Criterion} & \textbf{Statistical} & \textbf{Tree-Based} & \textbf{Neural} \\
\hline
Best for & Stable, low-volume series & Heterogeneous, feature-rich data & Large-scale, dense data \\
Intermittent demand & Croston variants & Strong (handles zeros well) & Sensitive to sparsity \\
Feature handling & Limited & Excellent & Requires engineering \\
Training time & Minutes & Minutes--hours & Hours--days \\
Interpretability & High & Medium (SHAP) & Low \\
Data requirements & Low & Medium & High \\
B\&M retail fit & Moderate & \textbf{Strong} & Weak--Moderate \\
\hline
\end{tabularx}
\caption{Practitioner Guidance: Model Selection by Use Case}
\end{table}

Beyond predictive accuracy, computational efficiency is another key factor used to evaluate forecasting models for practical use. Table 3 provides a comparison of training times in the whole-category setting for SAITS-imputed data. The results make it evident that ensemble approaches, especially XGBoost, are highly efficient: they require considerably less memory and train much faster than neural network approaches. For real-world applications where speed and accuracy matter, the lower computational footprint of the ensemble methods represents a clear advantage.

\subsection{Discussion}

The results offer several important insights into retail sales forecasting. Ensemble models such as LightGBM and XGBoost exhibit robust performance in both localized and aggregated configurations when they use nonimputed data, consistently achieving lower error rates while demonstrating significant training time advantages. This aligns with Borisov et al. \cite{Borisov}, who note that gradient-boosted decision tree ensembles often remain state-of-the-art for heterogeneous tabular data due to their robustness to data irregularities.

A key takeaway from our study is the importance of tailoring the modeling approach to the characteristics of the dataset. When data are segmented into individual groups and they do not share too much information between them, localized modeling can capture unique patterns more effectively. In our case, training on the whole category does not introduce enough inter-group information to bring improvements, a phenomenon that resonates with the observations by McElfresh et al. \cite{mcelfresh2024neuralnetsoutperformboosted}, who note that differences in dataset properties, such as skewed feature distributions and irregularities, can diminish the relative advantage of complex neural network architectures over simpler, well-tuned tree-based methods.

Furthermore, experiments with imputed versus nonimputed data underscore the critical role of data quality. Our analysis of SAITS imputation quality revealed that imputed values exhibit compressed variance and substantially different distributions from originals. Deep neural networks, as discussed by Ramesh and Usman \cite{Ramesh_Usman_2024}, tend to be more sensitive to inherent noise and missing values in tabular datasets. Our results extend this understanding by demonstrating that imputation-introduced artifacts also affect tree-based models, particularly the dramatic LightGBM performance collapse in Case C (RMSE increasing from 4.847 to 10.758). Detailed feature importance analysis revealed that while core predictors remain stable, the inclusion of imputed data shifts the relative importance of spatial features and introduces mild noise that degrades tree model performance.

While recent work by Zalando \cite{kunz2023deeplearningbasedforecasting} has demonstrated the presence of scaling laws for transformer-based models, where forecasting accuracy improves as training data size increases, such advantages were not observed in our experiments. This is likely due to the comparatively smaller scale of our dataset and the absence of centralized, high-density demand signals typical of e-commerce platforms. As a result, we did not observe a trade-off between computational cost and accuracy in favor of deep learning models. In our setting, ensemble methods remained both more efficient and more accurate, reinforcing their suitability for B\&M retail forecasting tasks characterized by fragmentation, intermittency, and limited historical coverage.

\section{Conclusion}

This study demonstrates that ensemble methods, particularly XGBoost and LightGBM, substantially outperform neural network architectures in B\&M retail forecasting when applied to localized, individual-group data. XGBoost achieved the best overall performance with an RMSE of 4.833, representing a significant improvement over naive baselines. Neural architectures exhibited sensitivity to intermittency and aggregation level; while SAITS-based imputation improved their performance in the whole-category setting, they still fell short of ensemble methods. Our imputation quality analysis revealed that SAITS introduces distributional compression and correlation shifts that particularly affect tree models in localized settings, explaining the dramatic LightGBM performance degradation in Case C.

For practitioners, these findings suggest that a localized modeling strategy leveraging tree-based ensembles ensures scalability, efficiency, and improved forecast accuracy, with direct applications in inventory management and demand planning. The superior computational efficiency of these methods makes them particularly suitable for real-time forecasting scenarios.

Several limitations warrant discussion. Aggregated error metrics may obscure performance variability across segments, and findings may not generalize to all retail environments. Future research should investigate hybrid approaches integrating tree-based efficiency with neural representation learning, and explore imputation methods that better preserve the underlying data structure. This work provides a systematic performance analysis under realistic conditions and offers actionable guidance for practitioners, reinforcing the importance of matching model complexity to problem characteristics.

\section{Statement on the Use of Artificial Intelligence}

The Large Language Models were used in order to improve phrasing and clarity of the text. All ideas, arguments, analysis, and conclusions are the original work of the authors.



\printbibliography

\end{document}